\documentclass{article}

\usepackage[T1]{fontenc}                
\usepackage{booktabs}                   
\usepackage{nicefrac}                   
\usepackage{microtype}                  
\usepackage{url}                        
\usepackage{xcolor}
\usepackage{xkcdcolors}

\usepackage{multirow}
\usepackage{wrapfig}
\usepackage{siunitx}
\usepackage{float}

\usepackage{pgfplots}

\usepackage[accepted]{icml2025}

\usepackage{caption}
\usepackage[pdfusetitle]{hyperref}
\hypersetup{
  pdfsubject={Proceedings of the International Conference on Machine Learning 2024},
  colorlinks=true,
  linkcolor=xkcdCrimson,
  urlcolor=xkcdBluish,
  citecolor=xkcdLightNavy,
}

\usepackage{bookmark}                   

\usepackage{amsmath}
\usepackage{amssymb}
\usepackage{amsfonts} 
\usepackage{amsthm}
\usepackage{mathtools}

\usepackage[shortcuts=abbr]{glossaries-extra}
\glsdisablehyper{}

\usepackage{amsmath,amsfonts,bm}

\def\eqref#1{equation~\ref{#1}}

\def\1{\bm{1}}

\def\vzero{{\bm{0}}}

\def\vc{{\bm{c}}}

\def\vm{{\bm{m}}}

\def\vq{{\bm{q}}}

\def\vu{{\bm{u}}}

\def\vx{{\bm{x}}}
\def\vy{{\bm{y}}}

\def\mA{{\bm{A}}}
\def\mB{{\bm{B}}}
\def\mC{{\bm{C}}}
\def\mD{{\bm{D}}}

\def\mI{{\bm{I}}}

\def\mK{{\bm{K}}}

\def\mP{{\bm{P}}}
\def\mQ{{\bm{Q}}}

\def\mPhi{{\bm{\Phi}}}

\def\mSigma{{\bm{\Sigma}}}

\DeclareMathAlphabet{\mathsfit}{\encodingdefault}{\sfdefault}{m}{sl}
\SetMathAlphabet{\mathsfit}{bold}{\encodingdefault}{\sfdefault}{bx}{n}

\def\emA{{A}}
\def\emB{{B}}

\newcommand{\R}{\mathbb{R}}

\usepackage[capitalise]{cleveref}

\usepackage{crossreftools}
\let\ORGhypersetup\hypersetup
\protected\def\hypersetup{\ORGhypersetup}
\usepackage[status=draft,layout=inline,author=TODO]{fixme}
\fxusetheme{color}

\theoremstyle{plain}

\theoremstyle{definition}

\theoremstyle{remark}

\newcommand\restrict[2]{{
  \left.\kern-\nulldelimiterspace 
  #1 
  \littletaller 
  \right|_{#2} 
  }}
\newcommand{\littletaller}{\mathchoice{\vphantom{\big|}}{}{}{}}

\DeclareMathSymbol{\shortminus}{\mathbin}{AMSa}{"39}
\newcommand{\pr}{\mathrm{p}}
\newcommand{\vbeta}{{\bm{\beta}}}
\newcommand{\mPsi}{{\bm{\Psi}}}
\newcommand{\mGamma}{{\bm{\Gamma}}}

\newcommand{\N}{\mathcal{N}}
\newcommand{\Nat}{\mathbb{N}}
\newcommand{\bigO}{\mathcal{O}}
\newcommand{\T}{^{\mathsf{T}}}

\newcommand\degree{D}
\newcommand\ncoefs{N}
\newcommand\nsteps{n}
\newcommand\ninputs{H}
\newcommand\nlatentchannels{M}
\newcommand\fhat{\hat{f}}
\newcommand\ft{f_{\le t}}
\newcommand\fthat{\hat{f}_{\le t}}

\newcommand\ftkhat{\hat{f}_{\le t_{k}}}
\newcommand\ftkpohat{\hat{f}_{\le t_{k + 1}}}
\newcommand{\diff}{\mathrm{d}}
\newcommand\ddt[1]{\frac{\diff{#1}}{\diff{t}}}
\newcommand\ddtau[1]{\frac{\diff{#1}}{\diff{\tau}}}
\newcommand\ddx[1]{\frac{\diff{#1}}{\diff{x}}}
\newcommand\legp{P}
\newcommand\normlegp{g}

\newcommand\Rplus{\R^{+}}
\newcommand\mAhippo{\mA_{\mathrm{H}}}
\newcommand\mAhippobar[1]{\bar{\mA}_{\mathrm{H}\if\relax\detokenize{#1}\relax\else,#1\fi}}
\newcommand\mBhippo{\mB_{\mathrm{H}}}
\newcommand\mBhippobar[1]{\bar{\mB}_{\mathrm{H}\if\relax\detokenize{#1}\relax\else,#1\fi}}
\newcommand\emAhippo[1]{\emA_{\mathrm{H,#1}}}
\newcommand\emBhippo[1]{\emB_{\mathrm{H,#1}}}
\newcommand\mArhippo{\mA_{\mathrm{R}}}

\newcommand\mArhippobar[1]{\bar{\mA}_{\mathrm{R,#1}}}

\newcommand\mAunhippobar[1]{\bar{\mA}_{\mathrm{U,#1}}}
\newcommand\mBunhippobar[1]{\bar{\mB}_{\mathrm{U,#1}}}

\newcommand{\defabbrcmd}[1]{
\expandafter\def\csname#1\endcsname{{\gls{#1}}}
}
\newcommand{\defabbrcmds}[1]{
\expandafter\def\csname#1\endcsname{{\gls{#1}}}
\expandafter\def\csname#1s\endcsname{{\glspl{#1}}}
}

\newabbreviation{ssm}{SSM}{state space model}
\defabbrcmds{ssm}
\newabbreviation{ode}{ODE}{ordinary differential equation}
\defabbrcmds{ode}
\newabbreviation{lds}{LDS}{linear dynamical system}
\defabbrcmds{lds}
\newabbreviation{lmu}{LMU}{Legendre Memory Unit}
\defabbrcmds{lmu}
\newabbreviation{hippo}{HiPPO}{high-order polynomial projection operator}
\defabbrcmd{hippo}
\newabbreviation{unhippo}{UnHiPPO}{uncertainty-aware HiPPO}
\defabbrcmd{unhippo}
\newabbreviation{unlssl}{UnLSSL}{uncertainty-aware LSSL}
\defabbrcmd{unlssl}
\newabbreviation{lssl}{LSSL}{Linear State Space Layer}
\defabbrcmd{lssl}

\begin{document}

\twocolumn[
\icmltitle{UnHiPPO: Uncertainty-aware Initialization for State Space Models}



\icmlsetsymbol{equal}{*}

\begin{icmlauthorlist}
  \icmlauthor{Marten Lienen}{tum,mdsi}
  \icmlauthor{Abdullah Saydemir}{tum}
  \icmlauthor{Stephan G\"unnemann}{tum,mdsi}
\end{icmlauthorlist}

\icmlaffiliation{mdsi}{Munich Data Science Institute, Technical University of Munich}
\icmlaffiliation{tum}{Department of Computer Science, Technical University of Munich}

\icmlcorrespondingauthor{Marten Lienen}{m.lienen@tum.de}

\icmlkeywords{generative, bayesian, inference, diffusion}

\vskip 0.3in
]

\setlength{\columnsep}{10pt}



\printAffiliationsAndNotice{}  

\begin{abstract}
  State space models are emerging as a dominant model class for sequence problems with many relying on the HiPPO framework to initialize their dynamics.
  However, HiPPO fundamentally assumes data to be noise-free; an assumption often violated in practice.
  We extend the HiPPO theory with measurement noise and derive an uncertainty-aware initialization for state space model dynamics.
  In our analysis, we interpret HiPPO as a linear stochastic control problem where the data enters as a noise-free control signal.
  We then reformulate the problem so that the data become noisy outputs of a latent system and arrive at an alternative dynamics initialization that infers the posterior of this latent system from the data without increasing runtime.
  Our experiments show that our initialization improves the resistance of state-space models to noise both at training and inference time.
\end{abstract}

\section{Introduction}\label{sec:introduction}

In \citeyear{voelker2019legendre}, \citeauthor{voelker2019legendre} proposed a novel memory cell, the \lmu{}, for recurrent neural networks.
Their cell keeps a continuously updated representation of input in a sliding window in terms of orthogonal polynomials.
After \citet{gu2020hippo} generalized \lmus{} to variable length input windows, they were further extended as \ssms{} by \citet{gu2021combining} and applied across a multitude of domains.
Their ability to capture temporal dependencies over long sequences made them a great fit for natural language processing \citep{mehta2022long}, time series analysis \citep{patro2024simba}, speech generation \citep{goel2022its} and more \citep{patro2024mamba360}.

A significant contribution to the success of \ssms{} is the \hippo{} initialization for their dynamics introduced by \citet{gu2020hippo}.
The \hippo{} dynamics compress a sequence into a finite-dimensional state that represents the projection of the sequence onto a basis of Legendre polynomials.
Importantly, maintaining this state requires only the current state and the next value in the sequence and is thus independent of the sequence length.
This enables the construction of \ssms{} that efficiently capture long-range dependencies with linear computational complexity.
Such \ssms{} are well-suited for deep learning applications where sequence length can be substantial, such as language \citep{gu2023mamba} or video modeling \citep{park2025videomamba}.

However, the original \hippo{} framework assumes that the observed data is noise-free -- an assumption that is easily violated.
In many practical scenarios, measurements are contaminated with noise due to sensor imperfections, environmental factors, or inherent variability in the data-generating process.
Thus, this noise-free assumption limits the applicability of \hippo{}-initialized \ssms{}, as they may perform suboptimally when exposed to noisy observations commonly encountered in real-world applications.

We address this limitation by extending the \hippo{} framework to explicitly account for measurement noise.
By reinterpreting \hippo{} as a linear stochastic control problem where data emerges as noisy observations of a latent system, we derive the \unhippo{} initialization for \ssms{}.
With \unhippo{}, an \ssm{} implicitly performs posterior inference in a linear dynamical system, making it robust against noise in the data without increasing computational complexity or changes to the model structure.

Our \textbf{contributions} can be summarized as follows:
\begin{itemize}
  \item We provide a thorough description and derivation of \hippo{} and its application to \ssm{} initialization in \cref{sec:legendre-polynomials,sec:hippo,sec:hippo-derivation}.
  \item In \cref{sec:control-to-inference}, we analyze \hippo{} with respect to its noise robustness using the framework of linear stochastic control theory.
        We modify \hippo{} based on this analysis so that it performs implicit posterior inference under the assumption of observation noise and propose a regularization technique to make this numerically stable.
  \item In \cref{sec:initialization}, we present our uncertainty-aware extension of \hippo{} and show how to apply it as an initialization in the \lssl{} model \citep{gu2021combining}.
        Furthermore, we analyze the effect of applying the time-varying \unhippo{} dynamics in a time-invariant way and why this can be interpreted as operating a model on different time scales.
  \item Our experiments in \cref{sec:experiments} demonstrate how the \unhippo{} initialization improves the robustness of \ssms{} against noise using the example of \lssl{}.
\end{itemize}

Find our implementation at \href{https://cs.cit.tum.de/daml/unhippo}{cs.cit.tum.de/daml/unhippo}.

\paragraph{Notation}
We use $0$-based indexing for the coefficient vectors and transition matrices to stay consistent with the natural numbering of polynomial basis functions where $g_{i}$ is of degree~$i$.
Furthermore, $t \in \Rplus$ denotes the current point in time and $\tau \in [0, t]$ another point in the signal up to time~$t$.
$x$ is a point in the subset of $\R$ where the polynomials under consideration form an orthogonal basis, e.g.\ $x \in [-1, 1]$ for Legendre polynomials.
$[\nsteps]$ refers to the set of integers $1, \ldots, \nsteps$.

\section{Legendre Polynomials}\label{sec:legendre-polynomials}

\begin{figure}[h]
  \centering
  \input{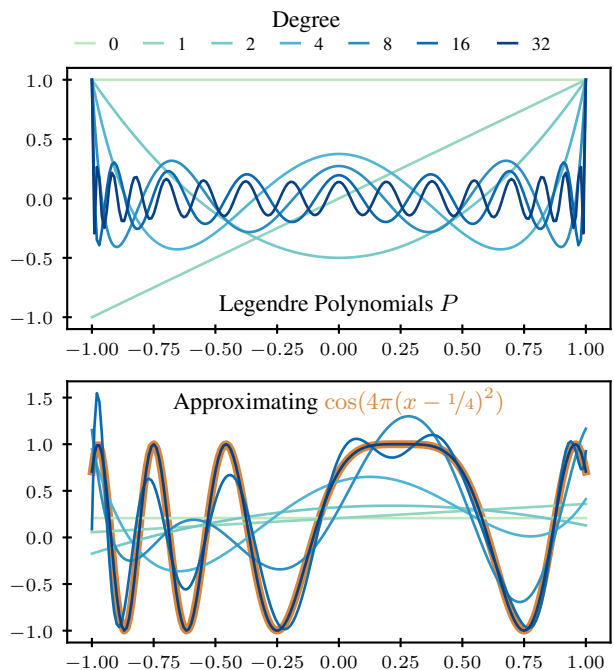}
  \caption{Function approximation with Legendre polynomials.}\label{fig:legendre}
\end{figure}

The Legendre polynomials $\legp_{i} : \R \to \R, i \in \Nat_{0}$ are a set of polynomials of increasing degree $i$ orthogonal with respect to the $L^{2}$ inner product $\langle f, g \rangle = \int_{-1}^{1} f(x)\,g(x)\,\diff{x}$.
Together, they form a basis of the space of $L^{2}$-integrable functions on $[-1, 1]$ and let us represent any such function~$f$ as
\begin{equation}
  \hat{f}(x) = \sum_{i = 0}^{\degree} \frac{\langle f, \legp_{i} \rangle}{{\|\legp_{i}\|}^{2}}\,\legp_{i}(x). \label{eq:legendre-projection}
\end{equation}
$\hat{f}$ is the best possible approximation to $f$ with respect to the distance induced by $\langle \cdot, \cdot \rangle$ in terms of polynomials up to degree $\degree$ and can be seen as the projection of $f$ onto the space spanned by the basis functions $\legp_{0}, \ldots, \legp_{\degree}$.
\cref{fig:legendre} illustrates these concepts.

In this sense, the vector $c_{i} = \langle f, \legp_{i} \rangle, i = 0, \ldots, \degree$ is the best possible compression of $f$ in the $L^{2}$ space into $\degree + 1$ numbers as it lets us reconstruct $f$ as $\hat{f}$.
For more background on polynomial approximation of signals in general and Legendre polynomials in particular, we refer the reader to \citep{vetterli2014foundations, arfken2008mathematical}.

\section{HiPPO}\label{sec:hippo}

Let's say, we observe a scalar signal $f(\tau) : \Rplus \to \R$ and want to process it in an online fashion with a downstream model $h : \R^{\ncoefs} \to \mathcal{X}$, for example time series classification or a running forecast.
Since $h$ processes fixed-length vectors, we need a vector representation of length $\ncoefs$ of the signal history $\ft \coloneq \restrict{f}{[0, t]}$ up to the current time~$t$ that is expressive and informative.

\citet{gu2020hippo} propose the \hippo{} that represents $\ft$ as the coefficient vector $\vc_{t}$ of the projection $\fthat$ of $\ft$ onto a basis of slightly modified Legendre polynomials $\normlegp_{i}$ up to degree~$\degree = \ncoefs - 1$.
While this representation is expressive, its computation via \cref{eq:legendre-projection} would require the complete signal $\ft$ at each point~$t$.
This would make both runtime and memory requirements scale with $t$, rendering this representation inefficient for long time series.

Crucially, \citet{gu2020hippo} show how the coefficients $\vc_{t_{k + 1}}$ can be computed approximately from coefficients $\vc_{t_{k}}$ at an earlier time~$t_{k}$ and the value of the signal $f$ at time~$t_{k + 1}$ with a linear update step.
This means that we only need to remember the current coefficients~$\vc_{t_{k}}$ instead of the complete signal history $\ft$ to update the compressed representation of the signal when a new observation $y_{t_{k+1}} = f(t_{k+1})$ comes in.
As a result, \hippo{} gives us $\ncoefs$ expressive features describing $\ft$ without having to store the complete history.
In fact, the memory requirements of \hippo{} are $\bigO(N)$ and thus independent of the length of the sequence.

In the following, we will present a condensed description of \hippo{} to introduce the formal background for our extension in \cref{sec:control-to-inference,sec:initialization}.
See \cref{sec:hippo-derivation} for a complete and expository derivation.

\paragraph{Shifting and scaling}
As a first step, the Legendre polynomials have to be shifted and scaled to form an orthonormal basis on the interval $[0, t]$. For that, \citet{gu2020hippo} choose the time-dependent inner product $\langle f, g \rangle_{t} = \nicefrac{1}{t} \int_{0}^{t} f(x)\,g(x)\,\diff{x}$, under which the basis polynomials
\begin{equation}
  \normlegp_{t,i}(\tau) = \sqrt{2i + 1}\,\legp_{i}(\phi_{t}(\tau)) \label{eq:norm-basis}
\end{equation}
are orthonormal. $\phi_{t}(\tau) = \nicefrac{2\tau}{t} - 1$ maps $[0, t]$ linearly onto $[-1, 1]$. This gives us the vector representation
\begin{equation}
  c_{t,i} = \langle \ft, \normlegp_{t,i} \rangle_{t} \label{eq:ft-projection}
\end{equation}
of the complete history of the signal $f$ until time $t$.

\paragraph{Online update}
To maintain the coefficient vector $\vc_{t}$ as time passes and we observe more of the signal $f$, \citet{gu2020hippo} take the time derivative of \cref{eq:ft-projection} to find the vector \ode{}
\begin{equation}
  \ddt{\vc_{t}} = -\frac{1}{t}\mAhippo \vc_{t} + \frac{1}{t}\mBhippo f(t), \label{eq:hippo-continuous}
\end{equation}
which describes the evolution of the coefficient vector $\vc_{t}$ in continuous time under the influence of the signal $f(t)$.
The \hippo{} matrix $\mAhippo$ and vector $\mBhippo$ are given by
\begin{equation}
  \begin{aligned}
    \emAhippo{ij}                  & = \begin{cases}
                                         \sqrt{2i + 1}\,\sqrt{2j + 1} & \mathrm{if}\ j < i, \\
                                         i + 1                        & \mathrm{if}\ j = i, \\
                                         0                            & \mathrm{if}\ j > i,
                                       \end{cases} \\
    \mathrm{and}\quad \emBhippo{i} & = \sqrt{2i + 1}. \label{eq:hippo-matrix}
  \end{aligned}
\end{equation}

\paragraph{Discretization} Applying the update to discrete data observed at times $t_{1}, t_{2}, \ldots$ requires a discrete analog of \cref{eq:hippo-continuous}.
To this end, \citet{gu2020hippo} discretize the equation into the recurrence
\begin{equation}
  \vc_{t_{k + 1}} = \bar{\mA}_{\mathrm{H}}\vc_{t_{k}} + \bar{\mB}_{\mathrm{H}}y_{t_{k + 1}}\label{eq:hippo-discrete}
\end{equation}
as described in \cref{sec:hippo-discretization}.
\cref{eq:hippo-discrete} can then serve as the basis for a recurrent neural network layer, similar to GRU \citep{cho2014learning} or LSTM \citep{hochreiter1997long}.

See \cref{sec:hippo-derivation} for an explanatory derivation of the results in this section.

\section{From Control to Inference}\label{sec:control-to-inference}

To understand the behavior of \hippo{} under measurement noise, we will analyze it as a linear stochastic control problem \citep{astrom1970introduction}.
In this theory, a continuous-discrete \lds{} with scalar control and observations is described as
\begin{align}
  \diff{\vx_{t}} & = \big( \mPhi \vx_{t} + \mGamma u_{t} \big)\diff{t} + \diff{\vbeta} \label{eq:lds-state-dynamics} \\
  y_{t_{k}}      & = \mPsi \vx_{t_{k}} + \varepsilon_{t_{k}}. \label{eq:lds-observation}
\end{align}
Such a system has a latent state $\vx_{t}$ that we can influence with a scalar control signal $u_{t}$ via its linear stochastic dynamics in \cref{eq:lds-state-dynamics}.
At discrete times~$t_{k}$, we take noisy measurements~$y_{t_{k}}$ of the system in \cref{eq:lds-observation}.
$\diff{\vbeta}$ is white noise, $\varepsilon_{t_{k}}$ is independent Gaussian noise and $\mPhi$, $\mGamma$ and $\mPsi$ are parameter matrices.
Note that \cref{eq:lds-state-dynamics,eq:lds-observation} include exactly two sources of noise: in the state dynamics and during the measurements.

If we compare the \hippo{} dynamics in \cref{eq:hippo-continuous} to the above system, we see that $\vc_{t}$ takes the role of the system state~$\vx_{t}$.
In contrast to what one might expect, the observed signal~$f(t)$ does not correspond to the observations $y_{t_{k}}$ in \hippo{} and, instead, it corresponds to the control signal~$u_{t}$, which is assumed to be noise-free in linear stochastic control theory.
This means that \hippo{} fundamentally assumes that there is no measurement noise on the signal~$f(t)$ nor any noise in the system dynamics themselves and that $\vc_{t}$ is a deterministic function of $\ft$.

However, in many applications, these assumptions do not hold.
This includes basically all time series modeling where the data is derived from a physical process such as a temperature, wind speed or power output, but also extends to exactly measurable but inherently noisy data such as user interactions and financial transaction logs.

Guarding the system against noise requires us to take it explicitly into account in the model.
Consequently, we design a continuous-discrete \lds{} based on the \hippo{} dynamics in \cref{eq:hippo-continuous} but with $f(t)$ modeled via noisy observations instead of a noise-free control signal.
In the end, $f(t)$ will not appear directly in the model at all.
Instead, we will infer the posterior distribution $\pr(\vc_{k} \mid y_{t_{1:k}})$, filtering out the effect of noise in the measurements and transitions.

To begin, we need to specify a model for $y_{t_{k}} = f(t_{k}) + \varepsilon$ linear in $c_{t_{k}}$.
Since $c_{t_{k}}$ represents the approximate signal $\ftkhat$ until $t_{k}$, we predict $f(t_{k})$ with
\begin{equation}
  \ftkhat(t_{k}) = \sum_{i = 0}^{\ncoefs} c_{t_{k},i}\,\normlegp_{t_{k},i}(t_{k}) = \mBhippo\T \vc_{t_{k}}, \label{eq:ftk-approximation}
\end{equation}
and model
\begin{equation}
  y_{t_{k}} = \mBhippo\T \vc_{t_{k}} + \varepsilon_{t_{k}} \quad \text{where} \quad \emBhippo{i} = \sqrt{2i + 1}.
\end{equation}

Secondly, we need to make the \hippo{} dynamics independent of $f(t)$ while keeping them linear.
For that, we substitute \cref{eq:ftk-approximation} for $f(t)$  in \cref{eq:hippo-continuous} and get
\begin{equation}
  \ddt{\vc_{t}} = \frac{1}{t} \big( \mBhippo \mBhippo\T - \mAhippo \big) \vc_{t}, \label{eq:hippo-data-free-dynamics}
\end{equation}
which we can simplify with the identity $\mBhippo \mBhippo\T - \mAhippo = \mAhippo\T - \mI$ which follows directly from the definitions in \cref{eq:hippo-matrix}.
Since the \hippo{} dynamics are exact, the dynamics in \cref{eq:hippo-data-free-dynamics} give us the exact evolution of $\vc_{t}$ if the true signal $\ft$ is just the polynomial $\fthat$.

By combining the model for $y_{t_{k}}$ and the data-free dynamics, we get the first version of our noise-resistant \lds{}.
\begin{equation}
  \begin{aligned}
    \diff{\vc_{t}} & = \frac{1}{t} \big( \mAhippo\T - \mI \big) \vc_{t}\diff{t} + \diff{\vbeta} \\
    y_{t_{k}}      & = \mBhippo\T \vc_{t_{k}} + \varepsilon_{t_{k}}
  \end{aligned} \label{eq:first-lds}
\end{equation}

\subsection{Extrapolation}\label{sec:extrapolation}

Inferring $\vc_{t}$ from data with \cref{eq:first-lds} will come down to two steps.
First, we solve \cref{eq:hippo-data-free-dynamics} forward in time from $t_{k}$ to $t_{k + 1}$ to get coefficients $\vc_{t_{k + 1}}$ that represent the extrapolation of $\ftkhat$ from $t_{k}$ to $t_{k + 1}$.
Second, we update the forecast $\vc_{t_{k + 1}}$ with the observed data $y_{t_{k+1}}$.
In contrast, \hippo{} uses the data directly in the forward solving of \cref{eq:hippo-continuous}.

A complication with our first step is that Legendre polynomials -- and by extension $\ftkhat$ -- are terribly unfit for extrapolation.
While each $\normlegp_{t,i}$ is well behaved within $[0, t]$, they diverge rapidly towards $\pm\infty$ outside of it -- like an $i$-th degree polynomial to be exact.

\begin{figure}[h]
  \centering
  \input{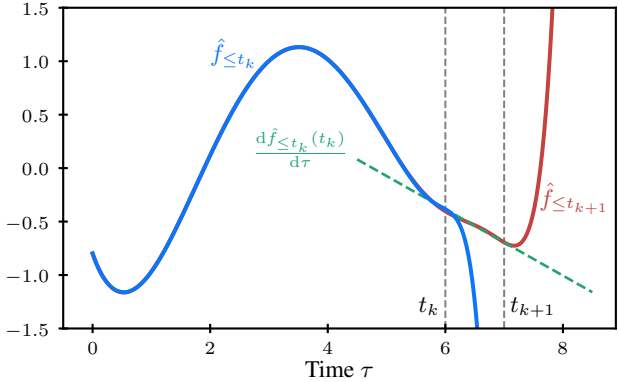}
  \caption{The Legendre polynomial $\ftkhat$ diverges rapidly beyond its domain $[0, t_{k}]$. If we extrapolate $\ftkhat$ with \cref{eq:hippo-data-free-dynamics}, $\ftkpohat$ will equal $\ftkhat$ but on the extended domain $[0, t_{k+1}]$. By regularizing the dynamics with the conditions in \cref{eq:regularization-conditions}, we can get the $\ftkpohat$ shown here, that extrapolates $\ftkhat$ linearly until $t_{k + 1}$.}\label{fig:extrapolation}
\end{figure}

This is a problem from two perspectives.
First, when we model $\ft$ with, for example, $\ncoefs = 256$ coefficients, $\ftkhat$ diverges like a degree $255$ polynomial outside of $[0, t_{k}]$ and any predicted $y_{t_{k + 1}}$ will almost surely be a horrible prediction of the true $f(t_{k + 1})$.
Second, numerical computations with a forecast of such magnitude will produce enough floating point cancellation to invalidate any results.

To reduce the order of divergence, we propose to regularize the system dynamics.
In particular, we are looking for dynamics that retain the shape of $\fhat$ as much as possible up to $t_{k}$ but evolve $\vc_{t}$ such that the slope of $\fhat$ remains roughly constant from $t_{k}$ onwards.
\cref{fig:extrapolation} visualizes this goal.
By retaining the shape of $\fhat$, we preserve the information embedded in $\vc_{t}$.

Formally, we can express our desired behavior of $\fthat$ as
\begin{equation}
  \ddt{\fthat(t)} = \ddtau{\fthat(t)} \quad \text{and} \quad \ddt{}\ddtau{\fthat(t)} = 0. \label{eq:regularization-conditions}
\end{equation}
In words, the first condition demands that the value at $t$ of the reconstruction $\fthat$, i.e.\ the current value of our signal, follow the gradient of $\fthat$ as we extend the domain and predict into the future.
If you compare to \cref{fig:extrapolation}, we want that the rightmost value of $\fthat$ follows the dashed line as we increase $t$.
The second condition says that this gradient shall not change.
Remember that the parameter of $\fthat$ is $\tau$, so the two derivatives in the conditions refer to the subscript and the parameter, respectively.

With \cref{eq:ftk-approximation}, we see immediately that
\begin{equation}
  \ddt{\fthat(t)} = \mBhippo\T\ddt{\vc_{t}}.
\end{equation}
The $\tau$-derivative of $\fthat(\tau)$ at $\tau = t$ becomes
\begin{equation}
  \ddtau{\fthat(t)} = \sum_{i = 0}^{\ncoefs} c_{t,i}\,\ddtau{\normlegp_{t,i}(t)} = \frac{2}{t}\,\mQ\T\vc_{t}
\end{equation}
where $Q_{i} = \sqrt{2i + 1}\ddx{\legp_{i}(1)} = \sqrt{2i + 1}\frac{i(i + 1)}{2}$ which can be derived from \cref{eq:norm-basis} and \citep[Eq. (12.24)]{arfken2008mathematical}
\begin{equation}
  \ddx{\legp_{i}(x)} = i\legp_{i - 1}(x) + x\ddx{\legp_{i - 1}(x)}.
\end{equation}

Now, we can combine \cref{eq:hippo-data-free-dynamics,eq:regularization-conditions} into the condition
\begin{equation}
  \begin{pmatrix}
    \mI \\ \mBhippo\T \\ \mQ\T
  \end{pmatrix}\ddt{\vc_{t}} = \frac{1}{t}\begin{pmatrix}
    \mAhippo\T - \mI \\ 2\mQ\T \\ \mQ\T
  \end{pmatrix}\vc_{t}.
\end{equation}
Since this is overdetermined, we cannot solve for $\ddt{\vc_{t}}$ exactly, but we can find an approximate solution with the pseudo-inverse denoted by $\dagger$,
\begin{equation}
  \ddt{\vc_{t}} = \frac{1}{t}\underbrace{{\begin{pmatrix}
          \mI \\ \mBhippo\T \\ \mQ\T
        \end{pmatrix}}^{\dagger}\begin{pmatrix}
      \mAhippo\T - \mI \\ 2\mQ\T \\ \mQ\T
    \end{pmatrix}}_{\eqcolon \mArhippo}\vc_{t}. \label{eq:regularized-dynamics}
\end{equation}
We call $\mArhippo$ the \emph{regularized \hippo{} matrix} and define the regularized \lds{}
\begin{equation}
  \begin{aligned}
    \diff{\vc_{t}} = \frac{1}{t}\mArhippo \vc_{t}\diff{t} + \diff{\vbeta}, \qquad y_{t_{k}} = \mBhippo\T \vc_{t_{k}} + \varepsilon_{t_{k}}. \label{eq:regularized-lds}
  \end{aligned}
\end{equation}

\subsection{Discretization}\label{sec:discretization}

At this point, we move from the semi-discretized to the fully discretized case.
As described by \citet[Section 6]{sarkka2019applied}, the discrete-time model equivalent to \cref{eq:regularized-lds} is
\begin{equation}
  \begin{aligned}
    \vc_{k + 1} & = \mArhippobar{k+1}\vc_{k} + \vq_{k} \\
    y_{k}       & = \mBhippo\T\vc_{k} + \varepsilon_{k}
  \end{aligned} \label{eq:discretized-lds}
\end{equation}
where $\mArhippobar{k+1}$ is the transition matrix from $t_{k}$ to $t_{k + 1}$, i.e.
\begin{equation}
  \mArhippobar{k+1} = \mI + \int_{t_{k}}^{t_{k + 1}} \frac{1}{\tau}\mArhippo\mArhippobar{k}\,\diff{\tau}. \label{eq:transition-matrix}
\end{equation}
$\vq_{k}$ and $\varepsilon_{k}$ are zero-mean Gaussian noise variables with covariance $\mSigma$ and $\sigma^{2}$, respectively, as hyperparameters.

In contrast to \hippo{}, the dynamics in \cref{eq:regularized-dynamics} do not depend on the observations.
This means that we can evaluate them, and in particular $\nicefrac{1}{t}\,\mArhippo$, at any $t$, not just the $t_{k}$ where we have observed data.
Therefore, we can use any \ode{} solver to approximate the transition matrix via \cref{eq:transition-matrix}.

We can, for example, approximate $\mArhippobar{k+1}$ with any integration rule considered by \citet{gu2021combining}, e.g.\ a forward Euler step $\mI + \frac{\Delta t}{t_{k}}\mArhippo$, a backward Euler step ${\big( \mI - \frac{\Delta t}{t_{k + 1}}\mArhippo \big)}^{-1}$ or the trapezoidal rule ${\big( \mI - \frac{\Delta t}{2t_{k + 1}}\mArhippo \big)}^{-1}\big( \mI + \frac{\Delta t}{2t_{k}}\mArhippo \big)$ where $\Delta t = t_{k + 1} - t_{k}$.
However, in contrast to \hippo{}, there exists an exact closed-form solution $\exp(\log(\nicefrac{t_{k + 1}}{t_{k}})\,\mArhippo)$ to \cref{eq:transition-matrix}, which we use instead of approximations.
We show in \cref{sec:discretization-visual} that it produces the most stable linear recurrence.
Note that it is a matrix exponential, not a component-wise application.

\subsection{Posterior Distribution}\label{sec:posterior}

Next, we connect the model specified in \cref{eq:discretized-lds} to the data.
Since both dynamics and observations are linear, we can compute the posterior $\pr(\vc_{k} \mid y_{1:k}) = \N(\vm_{k}, \mP_{k})$ in closed form via the Kalman filter \citep{kalman1960new, sarkka2019applied}.

For that, we put a standard Gaussian prior $\N(\vm_{0}, \mP_{0})$ on $\vc_{0}$, i.e.\ $\vm_{0} = \vzero$ and $\mP_{0} = \mI$.
Then, we can compute $\pr(\vc_{k} \mid y_{1:k}) = \N(\vm_{k}, \mP_{k})$ through the following recurrence.
First, we roll out the dynamics for one step to extend $\ftkhat$ to $\ftkpohat$.
\begin{equation}
  \begin{aligned}
    \vm_{k}^{-} & = \mArhippobar{k}\vm_{k-1} \\
    \mP_{k}^{-} & = \mArhippobar{k}\mP_{k-1}\mArhippobar{k}\T + \mSigma
  \end{aligned} \label{eq:kalman-predict}
\end{equation}
Then, we compare the prediction to the observed data $y_{k}$ and update the posterior parameters accordingly.
\begin{equation}
  \begin{aligned}
    v_{k}   & = y_{k} - \mBhippo\T\vm_{k}^{-} \\
    s_{k}   & = \mBhippo\T\mP_{k}^{-}\mBhippo + \sigma^{2} \\
    \mK_{k} & = \nicefrac{1}{s_{k}}\,\mP_{k}^{-}\mBhippo \\
    \vm_{k} & = \vm_{k}^{-} + \mK_{k}v_{k} \\
    \mP_{k} & = \mP_{k}^{-} - s_{k}\mK_{k}\mK_{k}\T
  \end{aligned} \label{eq:kalman-update}
\end{equation}
Since there is no $0$-th data point and thus no $t_{0}$, we set $t_{0} = t_{1}$ to compute $\mArhippobar{1}$, giving $\mArhippobar{1} = \mI$.

\paragraph{Numerical Stability}

Direct application of the Kalman filter can be numerically challenging, because the covariance matrix $\mP_{k}$ often has some eigenvalues close to zero.
The smallest eigenvalues can then flip into the negative due to floating point cancellation in the matrix difference in \cref{eq:kalman-update}, losing the positive semi-definiteness of $\mP_{k}$.
Furthermore, because floating point addition and multiplication are non-associative \citep{goldberg1991what}, $\mP_{k}$ can also lose its symmetry, making the filter diverge.
In our experiments, we solved the former issue by computing the Kalman filter in double precision and the latter with symmetrization $\mP_{k} \coloneq (\mP_{k} + \mP_{k}\T) / 2$ after each update.

\section{Uncertainty-aware Initialization}\label{sec:initialization}

Upon closer inspection of the posterior distribution of $\vc_{k}$, we notice two things.
First, the posterior covariance contains no information about the data as a consequence of the assumptions of the Kalman filter.
Second, we can combine the whole procedure in \cref{eq:kalman-predict,eq:kalman-update} into a single linear equation for the posterior mean $\vm_{k}$.
\begin{equation}
  \vm_{k} = \underbrace{\big( \mI - \mK_{k}\mBhippo\T \big)\mArhippobar{k}}_{\mAunhippobar{k}}\vm_{k-1} + \underbrace{\mK_{k}}_{\mBunhippobar{k}}y_{k} \label{eq:unhippo}
\end{equation}
We call $\mAunhippobar{k}$ and $\mBunhippobar{k}$ the \emph{uncertainty-aware \hippo{}} (UnHiPPO) \emph{matrix} and \emph{vector}, respectively.
\glsunset{unhippo}

\begin{figure}[h]
  \centering
  \includegraphics{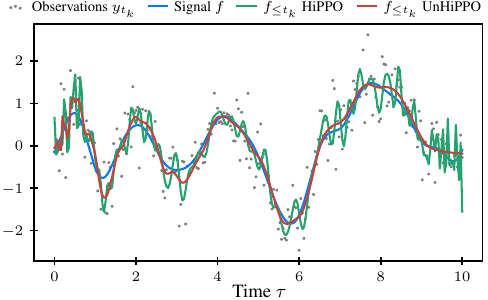}
  \caption{The uncertainty-aware \hippo{} dynamics filter out the majority of the noise.}\label{fig:comparison}
\end{figure}

\cref{eq:unhippo} is an uncertainty-aware equivalent to the discretized \hippo{} dynamics in \cref{eq:hippo-discrete}.
\cref{fig:comparison} compares the two in the face of noise.
It shows a random function sampled from a Gaussian process as the ground-truth signal and noisy observations derived from the signal by adding independent Gaussian noise.
On top of the data, we see the reconstructions $\ftkhat$ created from the coefficients gathered from the \hippo{} and \unhippo{} dynamics, respectively.
This lets us visually compare the effect of the observation noise on the features that these dynamics extract from the data.
We see that the noise introduces a lot of spurious high-frequency signal in the \hippo{} representation of the data, while the \unhippo{} dynamics filter out the majority of the noise and extract a close approximation of the ground-truth signal.

\begin{figure}[h]
  \centering
  \includegraphics{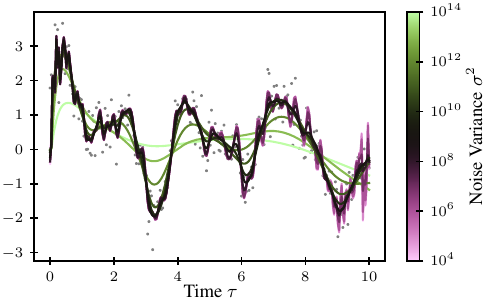}
  \caption{The $\sigma^{2}$ hyperparameter controls the level of filtering.}\label{fig:sigma-effect}
\end{figure}

The strength of the denoising is controlled by the $\sigma^{2}$ hyperparameter through its influence on the so-called innovation covariance $s_{k}$ in the Kalman update step in \cref{eq:kalman-update}.
\cref{fig:sigma-effect} visualizes the effect spectrum of $\sigma^{2}$.
Too small values mean that the signal is modeled closely including noise, while too large values make the dynamics ignore the data.
For optimal filtering, $\sigma^{2}$ needs to be adapted to the data.
Note the surprisingly large scale of $\sigma^{2}$ in \cref{fig:sigma-effect}.
This is necessary because $\sigma^{2}$ needs to overcome the $\mBhippo\T\mP_{k}^{-}\mBhippo$ term in $s_{k}$, which is large because of the magnitude of entries of the \hippo{} matrix in \cref{eq:hippo-matrix}.
An unfortunate side effect is that $\sigma^{2}$ cannot be interpreted as the noise variance of the data directly.

For the transition uncertainty, we always choose $\mSigma = \mI$.
In our experiments, it had no appreciable effect except for instability if it was very small.

\subsection{Uncertainty-aware Linear State Space Layer}\label{sec:lssl}

The \lssl{} \citep{gu2021combining} is a simple sequence model that maps a scalar input sequence $u_{k}, k \in [\nsteps]$ to a scalar output sequence $y_{k}$ via a latent state $\vc_{k} \in \R^{\ncoefs}$ and the recurrence
\begin{equation}
  \begin{aligned}
    \vc_{k} & = \mA\vc_{k-1} + \mB u_{k} \\
    y_{k}   & = \mC\vc_{k} + \mD u_{k}
  \end{aligned} \label{eq:lssl-dynamics}
\end{equation}
where $\mA \in \R^{\ncoefs \times \ncoefs}$, $\mB \in \R^{\ncoefs}$, $\mC \in \R^{1 \times \ncoefs}$ and $\mD \in \R$ are parameters.

\citet{gu2021combining} create a building block for neural sequence models from this by stacking $\ninputs$ independent copies of \cref{eq:lssl-dynamics} for an $\ninputs$-dimensional input, i.e.\ the output of the previous layer.
To improve the expressiveness of the system, they introduce $\nlatentchannels$ latent channels by increasing the dimension of the output $\vy_{k}$ to $\nlatentchannels$.
Since simple stacking means that each feature of the input $\vu_{k}$ is processed independently, they also introduce a position-wise map to mix the channels.
This map consists of a GELU nonlinearity followed by a linear layer that maps $\vy_{k} \in \R^{\ninputs \times \nlatentchannels}$ to the final output of dimensionality $\ninputs$, i.e.\ the weight matrix of the linear layer has dimensions $(\ninputs \cdot \nlatentchannels) \times \ninputs$.
The stacked parameters become $\mA \in \R^{\ninputs \times \ncoefs \times \ncoefs}$, $\mB \in \R^{\ninputs \times \ncoefs}$, $\mC \in \R^{\ninputs \times \nlatentchannels \times \ncoefs}$ and $\mD \in \R^{\ninputs \times \nlatentchannels}$.

The initialization of $\mA$ and $\mB$ is essential as \citet{gu2021combining} have shown and the \hippo{} initialization produced much stronger results than a random initialization.
They discretize the \hippo{} dynamics to
\begin{equation}
  \vc_{k} \approx \mAhippobar{k}\vc_{k-1} + \mBhippobar{k} y_{k} \label{eq:lssl-recurrence}
\end{equation}
where
\begin{equation}
  \mAhippobar{k} = {\bigg( \mI + \frac{\Delta t}{2t_{k}}\mAhippo \bigg)}^{\!\shortminus{}1}\bigg( \mI - \frac{\Delta t}{2t_{k}}\mAhippo \bigg)
\end{equation}
and
\begin{equation}
  \mBhippobar{k} = {\bigg( \mI + \frac{\Delta t}{2t_{k}}\mAhippo \bigg)}^{\!\shortminus{}1}\frac{\Delta t}{t_{k}}\mBhippo.
\end{equation}
See \cref{sec:hippo-discretization} for a derivation of this discretization.
Then, they initialize $\mA_{i} = \mAhippobar{i}$ and $\mB_{i} = \mBhippobar{i}$ where they set $\Delta t = 1$ and vary $t_{k}$ log-uniformly between $t_{\mathrm{min}}$ for $i = 1$ and $t_{\mathrm{max}}$ for $i = \ninputs$.

We adapt this into the \unlssl{} by initializing
\begin{equation}
  \mA_{i} = \mAunhippobar{\lfloor t \rfloor} \qquad \text{and} \qquad \mB_{i} = \mBunhippobar{\lfloor t \rfloor}
\end{equation}
where $t$ varies in the same way as for \lssl{}.
One difference to \lssl{} is in how these initializations are computed.
In contrast to \lssl{} where we can get the discretized dynamics at any $t$ directly, for \unlssl{} we compute them for all integer steps $t \in [t_{\mathrm{max}}]$ and then select a subset.
In theory, we could jump to any $t$ directly in the Kalman update in \cref{eq:kalman-update}, but that would increase the uncertainty $\mP_{k}^{-}$ as if there was no data before $t$, changing the dynamics.
Instead, we also compute all intermediate steps, which mirrors the more realistic setting where we also observe data at $1, 2, \ldots, t - 1$.
Note that this only happens once for initialization and has negligible runtime cost.

\citet{gu2021combining} improve the runtime of \lssl{} by rewriting \cref{eq:lssl-recurrence} from a recurrence to a convolution with a Krylov kernel that can be constructed in $O(\log \nsteps)$ time.
Since \unlssl{} is just a different initialization for the same dynamics, the same optimization applies.
Note that the Krylov kernel for \unlssl{} needs to be constructed in double precision, though it can be cast to single precision before the convolution.

$\mA_{i}$ and $\mB_{i}$ can be fine-tuned with gradient descent during model training.
However, \citet{gu2021combining} have shown that training $\mA_{i}$ and $\mB_{i}$ has only a minor effect on \lssl{} performance in sequence classification, but prevents caching the Krylov kernel and increases training time significantly.
Therefore, we keep the \ssm{} parameters fixed in our experiments.

\begin{figure}[h]
  \centering
  \includegraphics{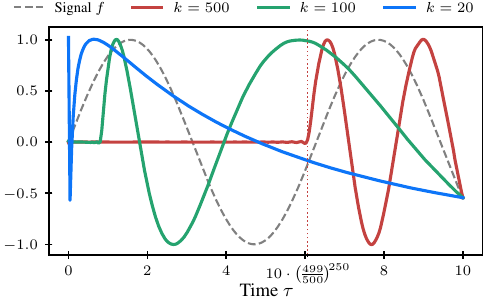}
  \caption{Reconstruction $\ftkhat$ of signal $f$ obtained through repeated application of $\bar{\mA}_{k}$ and $\bar{\mB}_{k}$ for fixed $k$. Data observed at 250 equispaced locations between 0 and 10. Each application of $\bar{\mA}_{k}$ compresses the reconstruction $\ftkhat$ by $\nicefrac{k-1}{k}$.}\label{fig:time-invariance}
\end{figure}

\paragraph{Time-invariant Dynamics}
Both the \unhippo{} dynamics in \cref{eq:unhippo} and the \hippo{} dynamics in \cref{eq:lssl-dynamics} are time-varying.
However, the \lssl{} and \unlssl{} layers fix the dynamics at a point in time and then apply them repeatedly.
We can understand the effect of applying the dynamics in a time-invariant way visually from \cref{fig:time-invariance}.

Semantically, the dynamics $\bar{\mA}_{k+1}$ of either \hippo{} or \unhippo{} take the reconstruction $\ftkhat$ of the signal from $0$ to $t_{k}$ represented by $\vc_{k}$ and extend it by one step.
Then, $\bar{\mB}_{k+1}$ updates the extended reconstruction, so that it fits the data at $t_{k+1}$.
To understand the effect of repeated application of the same $\bar{\mA}_{k+1}$, it is instructive to imagine what happens if we map the signal back to the domain $[-1, 1]$ of the Legendre polynomials with $\phi_{t}$ from \cref{sec:hippo}.
In that view, $\bar{\mA}_{k+1}$ squeezes the data that is encoded in $\ftkhat$ as a function on $[-1, 1]$ into $[-1, 1 - \nicefrac{2}{k+1}]$ and $\bar{\mB}_{k+1}$ fills the freed up interval $[1 - \nicefrac{2}{k+1}, 1]$ with the new data.
So, repeated application of $\bar{A}_{k+1}$ squeezes the information encoded in $\ftkhat$ by $\nicefrac{k}{k+1}$ each time.
\cref{fig:time-invariance} shows this as the 250-fold application of $\bar{\mA}_{500}$ has compressed the initial value of $\ftkhat = 0$ exactly by a factor of ${(\nicefrac{499}{500})}^{250}$.
The fact that the $\ftkhat$ for large $k$ contain the whole signal while for small $k$ they contain mostly the recent past, illuminates why initializing $\mA$ and $\mB$ with $\bar{\mA}_{k}$ and $\bar{\mB}_{k}$ for various $k$ in \lssl{} and \unlssl{} lets the model take multiple timescales into account as \citet{gu2021combining} report.

\section{Experiments}\label{sec:experiments}

We experiment with a multi-layer \lssl{} and \unlssl{} architecture with linear encoder and decoder as described by \citet{gu2021combining}.
\cref{sec:experiment-details} gives further details on the hyperparameters.

\begin{figure}
  \centering
  \includegraphics{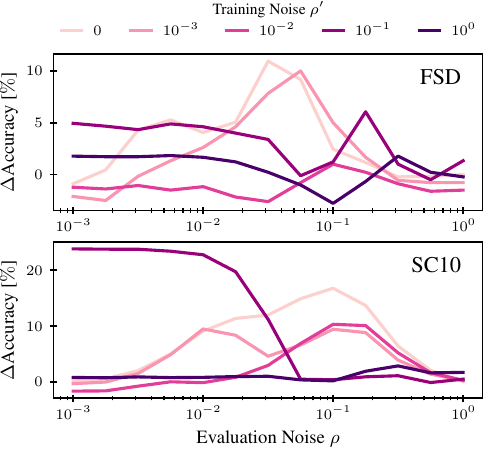}
  \caption{Accuracy difference between \unlssl{} and \lssl{} in speech classification trained on a noise level $\rho'$ and evaluated on a range of noise levels $\rho$.}\label{fig:delta}
\end{figure}
We evaluate \unlssl{} on two sequence classification datasets, the Free Spoken Digits dataset (FSD) \citep{jackson2018freespokendigitdataset} and a 10-class subset of the Speech Commands dataset (SC10) \citep{warden2018speech}.
FSD contains \num{3000} recordings of \num{6} speakers pronouncing each digit from 0 to 9 at a sample rate of \num{8000}.
We loop recordings shorter than one second and then cut them so that each sample is a univariate sequence of length \num{8000}.
SC10 is a multi-speaker dataset consisting of one second clips each being a recording of one spoken word.
Each sample is a univariate sequence of \num{16000} steps representing the raw waveform of the audio.

For our experiments, where we measure the effect of noise in the data, we add independent Gaussian noise $\delta \sim \N(0, \rho^{2})$ to the training and/or test data.

\paragraph{Noise Robustness}
For \cref{fig:delta}, we trained \lssl{} and \unlssl{} on data with varying noise levels $\rho'$ and evaluated the classification accuracy on a test set across a spectrum of noise levels $\rho$.
The results show that the \unhippo{} initialization improves the model's robustness against other noise levels than it was trained on, both lower and higher.

\begin{figure}
  \centering
  \includegraphics{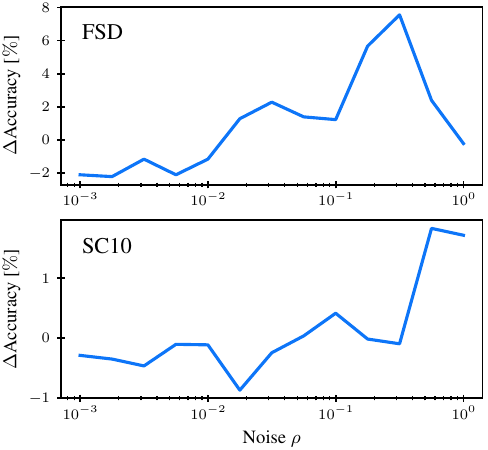}
  \caption{Accuracy difference between \unlssl{} and \lssl{} in speech classification for models trained and evaluated on a range of noise levels $\rho$.}\label{fig:accuracy}
\end{figure}
\paragraph{Training Noise}
In the real world, training and evaluation data often have similar levels of noise.
To investigate this case, we trained \lssl{} and \unlssl{} on a range of noise levels $\rho$ and then evaluated their accuracy on the test at the same noise level.
\cref{fig:accuracy} shows that the \unhippo{} initialization hinders the model's performance slightly at low noise levels while improving it as the noise level in the data increases.

\begin{figure}
  \centering
  \includegraphics{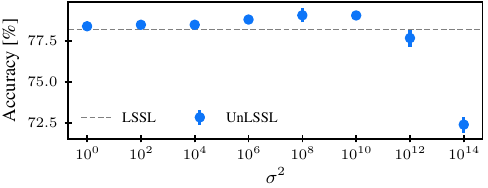}
  \caption{The $\sigma^{2}$ parameter adapts the noise filtering to the level of noise in the data on SC10 with $\rho = 0.1$. Error bars denote 1 standard deviation.}\label{fig:sigma2}
\end{figure}
\paragraph{Effect of $\sigma^{2}$}
$\sigma^{2}$ in the Kalman filter in \cref{eq:kalman-update} influences how much noise the \unhippo{} dynamics filter out.
This lets the user adapt \unhippo{} to the expected level of noise in the data.
See \cref{fig:sigma2} for an evaluation where we have trained \unlssl{} on SC10 with Gaussian noise of $\rho = 10^{-1}$ and varied the $\sigma^{2}$ parameter.
The results show that \unlssl{} achieves a better fit on noisy data than \lssl{}.

\paragraph{Discretization Methods}
While \lssl{} relies on a trapezoidal-like discretization for the \hippo{} matrix,
\begin{wraptable}[11]{r}{2in}
  \centering
  \vspace{-\columnsep}
  \caption{Time to compute discretized $\mArhippobar{k}$ for $\ncoefs = 256$ and resulting \unlssl{} accuracy on SC10 with $\rho = 10^{-1}$ noise.}\label{tab:discretizations}
  \begin{tabular}{rcc}
    \toprule
    Method      & Time                    & Accuracy \\
    \midrule
    closed-form & \SI{2.8}{\milli\second} & 79.3\%   \\
    trapezoidal & \SI{1.8}{\milli\second} & -        \\
    forward     & \SI{0.1}{\milli\second} & -        \\
    backward    & \SI{1.6}{\milli\second} & 79.8\%   \\
    \bottomrule
  \end{tabular}
\end{wraptable}
we use the closed-form solution in terms of the matrix exponential for the \unhippo{} matrix in \cref{eq:transition-matrix}.
\cref{tab:discretizations} shows that the closed-form solution is only 50\% slower than the trapezoidal approximation, which is negligible since this only needs to be computed once to initialize the \ssm{}.
An \unlssl{} model initialized with the resulting \unhippo{} matrices reaches comparable performance for the closed-form solution and a stable backwards Euler approximation.
Both the trapezoidal and forward Euler approximations produce an \ssm{} initialization that diverges on the long sequences of SC10.
\cref{sec:discretization-visual} has a visualization of the effect of the discretization methods on the \hippo{} and \unhippo{} matrices.

\section{Related Work}\label{sec:related-work}

\citet{yu2023robustifying} introduce a diagonalization of the \hippo{} matrix through an approach called ``perturb-then-diagonalize''.
Their parametrization improves runtime performance over a non-diagonal one and is more robust against noise than the diagonalizations proposed by \citet{smith2023simplified,gu2022parameterization}.
\citet{yu2025hope} propose an alternative parametrization for \ssms{} detached from \hippo{} and instead based on Hankel operator theory.
Their HOPE scheme improves the training stability of \ssms{} and makes them robust against noise appended to the end of a sequence.
However, they do not evaluate noise on the data itself.
\citet{agarwal2023spectral} derive an \ssm{} from the spectral filtering algorithm \citep{hazan2017learning} which is based on an \lds{} with noisy outputs similar to our setup in \cref{sec:control-to-inference}.
\citet{zhou2024kalmanssm} describe an \ssm{} that parametrizes a Kalman filter with a learnable dynamics matrix.
They accelerate their model by rewriting the recurrence as a convolution with a Krylov kernel similar to \citep{gu2021combining}.
However, they do not connect it to \hippo{} theory or \ssm{} initialization in general.

\citet{gu2022how} present additional background on the mechanism behind \hippo{}. \citet{gupta2022diagonal} explore a simplified alternative to \hippo{} with a diagonal transition matrix.
\citet{liu2025autocorrelation} analyze the role of initializations in state space models discretized with the zero-order hold approximation.

\section{Limitations}\label{sec:limitations}

\citet{gu2022efficiently} have proposed an alternative parametrization of the \hippo{} matrix $\mAhippo$ in terms of a diagonal and a low-rank matrix, which they call a \emph{structured} \ssm{}.
With this parametrization, the matrix multiplication in the \ssm{} recurrence can be computed in $O(\ncoefs)$ instead of $O(\ncoefs^{2})$.
A similarly structured parametrization of either the regularized \hippo{} matrix $\mArhippo$ in \cref{eq:regularized-dynamics} or the uncertainty-aware \hippo{} matrix $\mAunhippobar{}$ in \cref{eq:unhippo} eluded us, unfortunately, because of the pseudo-inverse and Kalman filter equations, respectively.

\section{Conclusion}\label{sec:conclusion}

We analyzed \hippo{} as a linear stochastic control problem and understood why it is not robust against noise.
By rewriting the control problem as a system controlled by a latent state with noisy observations, we were able to derive the \unhippo{} dynamics that perform posterior inference in that system and thereby filter out noise.
The resulting \unhippo{} initialization for \ssms{} offers a configurable amount of noise filtering without any modifications to the model structure or runtime costs.
We have furthermore shown in our experiments how such built-in noise filtering can improve the robustness of \ssms{} against noise in the real world.

\section*{Acknowledgments}

ML was funded by the Bavarian State Ministry for Science and the Arts within the framework of the Geothermal Alliance Bavaria project.
We want to thank the Munich Center for Machine Learning for providing compute resources.
Furthermore, we are thankful to Marcel Kollovieh for his insightful feedback and his support with \cref{fig:extrapolation}.

\section*{Authors' Contributions}

ML conceived the project idea, developed the theory, implemented the UnHiPPO initialization and wrote the manuscript.
AS implemented the majority of the pipeline, designed and executed the empirical evaluation (i.e.\ selected the datasets and implemented model training and data loading to generate the quantitative results for visualization and analysis), identified a technical correction in the mathematical formulation, and  corrected (Un)LSSL implementations.
SG contributed to the design of the method and provided scientific guidance.
All authors reviewed and discussed results, both theoretical and empirical, throughout the project.

\section*{Software}

For our results, we rely on excellent software packages, notably \texttt{numpy} \citep{harris2020array}, \texttt{scipy} \citep{virtanen2020scipy}, \texttt{pytorch} \citep{paszke2019pytorch}, \texttt{einops} \citep{rogozhnikov2021einops}, \texttt{matplotlib} \citep{hunter2007matplotlib}, \texttt{hydra} \citep{yadan2019hydra} and \texttt{jupyter} \citep{granger2021jupyter}.

\section*{Impact Statement}

This paper presents work whose goal is to advance the field of Machine Learning.
There are many potential societal consequences of our work, none of which we feel must be specifically highlighted here.

\bibliography{unssm}
\bibliographystyle{icml2025}

\clearpage
\onecolumn
\appendix

\bookmarksetupnext{level=part}
\pdfbookmark{Appendix}{appendix}

\section{HiPPO Derivation}\label{sec:hippo-derivation}

A complete derivation can, of course, also be found in the original work of \citep{gu2020hippo}.
However, the original derivation includes several additional degrees of freedom to support multiple \hippo{} variants.
Since the variant based on Legendre polynomials is the only one relevant to our work and also the one most relevant to follow-up works \citep{gu2021combining,gu2022efficiently}, we give a specialized derivation here that eschews unnecessary symbols.

\subsection{Choice of Basis}\label{sec:choice-of-basis}

Let's consider the case where we observe a continuous scalar signal $f(\tau) : \Rplus \to \R$ and at any point~$t$ in time we want to encode the history $\ft \coloneq \restrict{f}{[0, t]}$ of the signal that we have observed so far into an $\ncoefs$ dimensional vector $\vc_{t}$.
We consider $\vc_{t}$ a good representation of $\ft$, if we can turn it back into an approximate version $\fthat$ of $\ft$, we can rely on the polynomial approximation framework for signal processing \citep{vetterli2014foundations}.
In this framework, we begin by choosing an inner product $\langle \cdot, \cdot \rangle$ between functions on $[0, t]$ to induce a distance via $d(f, g) = \sqrt{\|f - g\|}$ and $\|f - g\| = \langle f - g, f - g \rangle$ and thus define exactly what $\fthat \approx \ft$ should mean.
Equipped with this inner product, we can then derive a set of $\ncoefs$ orthonormal polynomials $\normlegp_{i}, i = 0, \ldots, \ncoefs - 1$ on $[0, t]$ to form the basis of a finite-dimensional function space in which we can approximate the continuous signal~$\ft$.
Finally, we can find $\vc_{t}$ by projecting $\ft$ onto $\normlegp_{i}$, i.e.\ $c_{t,i} = \langle \ft, \normlegp_{i} \rangle$.

Our derivation of the Legendre-based \hippo{} begins with an inner product between two functions $f$ and $g$ on $[0, t]$.
We will choose
\begin{equation}
  {\langle f, g \rangle}_{t} = \frac{1}{t} \int_{0}^{t}f(\tau)\,g(\tau)\,\diff{\tau}, \label{eq:hippo-inner-product}
\end{equation}
which is just the $L^{2}$ inner product between functions except for a constant factor $\nicefrac{1}{t}$.
While the constant factor has no effect on orthogonality, it will be useful later when we update the representation $\vc_{t}$ with new data.
In terms of function approximation, this inner product induces the distance
\begin{equation}
  {\big(d(f, g)\big)}^{2} = \frac{1}{t} \int_{0}^{t} {(f(\tau) - g(\tau))}^{2}\,\diff{\tau}.
\end{equation}
This means that the approximation $\fthat$ we are looking for minimizes the squared difference to $\ft$ across the segment from $0$ to $t$ with equal weight given to each $\tau$.

Next, we need the polynomial basis.
Since the Legendre polynomials $\legp_{i}$ are an orthogonal basis on $[-1, 1]$ for the $L^{2}$ inner product, they are also orthogonal on $[0, t]$ with respect to our shifted and scaled inner product if we shift and scale them linearly from $[-1, 1]$ to $[0, t]$.
For this shifting, we define $\phi_{t}(\tau) = \nicefrac{2\tau}{t} - 1$ that takes $\tau \in [0, t]$ to $x \in [-1, 1]$.
This gives us $\legp_{i}(\phi_{t}(\tau))$ as an orthogonal basis on $[0, t]$.
However, the projection of $\ft$ onto the basis is simplified if the basis is orthonormal, so we need to compute the normalization constants of this shifted basis.
With $\phi_{t}'(\tau) = \nicefrac{2}{t}$ and substitution integration, we get
\begin{equation}
  {\|\legp_{i}(\phi_{t}(\tau))\|}_{t}^{2} = {\langle \legp_{i}(\phi_{t}(\tau)), \legp_{i}(\phi_{t}(\tau)) \rangle}_{t} = \frac{1}{t} \int_{0}^{t} {\legp_{i}(\phi_{t}(\tau))}^{2}\,\diff{\tau} = \frac{1}{2} \int_{-1}^{1} {\legp_{i}(x)}^{2}\,\diff{x} = \frac{1}{2i + 1}.
\end{equation}
The last step uses the normalization constant of the unscaled Legendre polynomials of $\nicefrac{2}{2i + 1}$.
This gives us
\begin{equation}
  \normlegp_{t,i}(\tau) = \sqrt{2i + 1}\,\legp_{i}(\phi_{t}(\tau)), \quad i = 0, \ldots, \degree
\end{equation}
as a basis of orthonormal polynomials on $[0, t]$ up to any degree~$\degree$.

Computing the representation $\vc_{t} \in \R^{\ncoefs}$ thus means evaluating $c_{t,i} = {\langle \ft, \normlegp_{t,i} \rangle}_{t}$ with \cref{eq:hippo-inner-product} for $i = 0, \ldots, \ncoefs - 1$.

\subsection{Linear Dynamics}\label{sec:linear-dynamics}

How does the representation $\vc_{t}$ of $\ft$ change as we observe more of the signal?
In the continuous case, the answer to this question is the \ode{}
\begin{equation}
  \ddt{c_{t,i}} = \ddt{} {\langle \ft, \normlegp_{t,i} \rangle}_{t} = \ddt{} \frac{1}{t} \int_{0}^{t}f(\tau)\,\normlegp_{t,i}(\tau)\,\diff{\tau}= -\frac{1}{t^{2}} \int_{0}^{t}f(\tau)\,\normlegp_{t,i}(\tau)\,\diff{\tau} + \frac{1}{t} \ddt{}  \int_{0}^{t}f(\tau)\,\normlegp_{t,i}(\tau)\,\diff{\tau}. \label{eq:hippo-deriv-ct-form-1}
\end{equation}
The first part is just $-\nicefrac{1}{t}\,c_{t,i}$ but the derivative of the integral in the second part needs some more work.
We can exchange differentiation and integration with Leibniz's rule and get
\begin{equation}
  \ddt{}  \int_{0}^{t}f(\tau)\,\normlegp_{t,i}(\tau)\,\diff{\tau} = f(t)\,\normlegp_{t,i}(t) + \int_{0}^{t}f(\tau)\,\ddt{\normlegp_{t,i}(\tau)}\,\diff{\tau}. \label{eq:hippo-deriv-intermediate-form-1}
\end{equation}
To progress, we need to use some properties of Legendre polynomials.
For the first part, we use that $\legp_{i}(1) = 1$ for all $i$ and therefore $\normlegp_{t,i}(t) = \sqrt{2i + 1}$.
For the second part, we use the identity (see \citep[Eq. (8)]{gu2020hippo} and \citep[Eqs. (12.23) and (12.24)]{arfken2008mathematical})
\begin{equation}
  (x + 1)\ \ddx{}P_{i}(x) = iP_{i}(x) + \sum_{j = 1}^{i} (2(i - j) + 1)\,P_{i-j}(x) \label{eq:legendre-deriv-identity}
\end{equation}
to derive $\ddt{}g_{t,i}(\tau)$.
Note that
\begin{equation}
  \ddt{}\phi_{t}(\tau) = \ddt{}\bigg[\frac{2\tau}{t} - 1\bigg] = -\frac{2\tau}{t^{2}} = -\frac{1}{t}\bigg(\frac{2\tau}{t} - 1 + 1\bigg) = -\frac{1}{t} \big(\phi_{t}(\tau) + 1\big). \label{eq:phi-deriv-identity}
\end{equation}
With \cref{eq:legendre-deriv-identity,eq:phi-deriv-identity} and recognizing $\phi_{t}(\tau)$ as $x$, we get
\begin{equation}
  \begin{aligned}
    \ddt{}\normlegp_{t,i}(\tau) & = \sqrt{2i + 1}\ \ddt{}\legp_{i}(\phi_{t}(\tau)) \\
                                & = \sqrt{2i + 1}\ \ddt{\phi_{t}(\tau)}\,\frac{\partial}{\partial t}\legp_{i}(\phi_{t}(\tau)) \\
                                & = -\frac{1}{t} \sqrt{2i + 1}\,\big(\phi_{t}(\tau) + 1\big)\,\frac{\partial}{\partial t}\legp_{i}(\phi_{t}(\tau)) \\
                                & = -\frac{1}{t} \sqrt{2i + 1}\,\bigg[ iP_{i}(\phi_{t}(\tau)) + \sum_{j = 1}^{i} (2(i - j) + 1)\,P_{i-j}(\phi_{t}(\tau)) \bigg] \\
                                & = -\frac{1}{t} \bigg[ i\normlegp_{t,i}(\tau) + \sqrt{2i + 1}\,\sum_{j = 1}^{i} \sqrt{2(i - j) + 1}\,\normlegp_{t,i-j}(\tau) \bigg].
  \end{aligned}
\end{equation}
When we plug this into the second part of \cref{eq:hippo-deriv-intermediate-form-1}, we see that
\begin{equation}
  \begin{aligned}
    \int_{0}^{t}f(\tau)\,\ddt{\normlegp_{t,i}(\tau)}\,\diff{\tau} & = -\frac{1}{t} \int_{0}^{t}f(\tau)\,\bigg[ i\normlegp_{t,i}(\tau) + \sqrt{2i + 1}\,\sum_{j = 1}^{i} \sqrt{2(i - j) + 1}\,\normlegp_{t,i-j}(\tau) \bigg]\,\diff{\tau} \\
                                                                  & = -\bigg[ i\,\underbrace{{\langle f, \normlegp_{t,i} \rangle}_{t}}_{c_{t,i}} + \sqrt{2i + 1}\,\sum_{j = 1}^{i} \sqrt{2(i - j) + 1}\,\underbrace{{\langle f, \normlegp_{t,i-j} \rangle}_{t}}_{c_{t,i-j}} \bigg] \label{eq:hipp-deriv-intermediate-form-2}
  \end{aligned}
\end{equation}
If we plug \cref{eq:hipp-deriv-intermediate-form-2} into \cref{eq:hippo-deriv-intermediate-form-1} and that in turn into \cref{eq:hippo-deriv-ct-form-1}, we see that
\begin{equation}
  \begin{aligned}
    \ddt{c_{t,i}} & = -\frac{1}{t^{2}} \int_{0}^{t}f(\tau)\,\normlegp_{t,i}(\tau)\,\diff{\tau} + \frac{1}{t} \ddt{}  \int_{0}^{t}f(\tau)\,\normlegp_{t,i}(\tau)\,\diff{\tau} \\
                  & = -\frac{1}{t} c_{t,i} + \frac{1}{t} \bigg[ \sqrt{2i + 1}f(t) - \Big[ i\,c_{t,i} + \sqrt{2i + 1}\,\sum_{j = 1}^{i} \sqrt{2(i - j) + 1}\,c_{t,i-j} \Big]  \bigg]. \label{eq:hippo-deriv-ct-elementwise}
  \end{aligned}
\end{equation}
Now we can separate terms involving $\vc$ and $f$ and write the $\ncoefs$ instances of \cref{eq:hippo-deriv-ct-elementwise} for the $\ncoefs$ components of $\vc$ as a single vector differential equation.
This gives us the full continuous coefficient dynamics from \citep{gu2020hippo}
\begin{equation}
  \ddt{\vc_{t}} = -\frac{1}{t} \mAhippo \vc_{t} + \frac{1}{t} \mBhippo f(t)  \label{eq:hippo-dynamics}
\end{equation}
with the \hippo{} matrix and vector
\begin{equation}
  \emAhippo{ij} = \begin{cases}
    \sqrt{2i + 1}\,\sqrt{2j + 1} & \text{if}\ j < i \\
    i + 1                        & \text{if}\ j = i \\
    0                            & \text{if}\ j > i \\
  \end{cases} \qquad \text{and} \qquad \emBhippo{i} = \sqrt{2i + 1}, \label{eq:hippo-matrix-appendix}
\end{equation}
respectively.

\subsection{Discretization}\label{sec:hippo-discretization}

Usually, we observe the signal~$f$ at discrete time points~$t_{k}$ instead of truly continuously.
Therefore, we cannot evaluate the exact, continuous dynamics in \cref{eq:hippo-dynamics} and have to discretize them instead, so that we only ever need to evaluate $f$ at the discrete observation points $t_{k}$.

With a continuous signal, we would compute $\vc_{t_{k + 1}}$ from $c_{t_{k}}$ by integrating \cref{eq:hippo-dynamics} from $t_{k}$ to $t_{k + 1}$, i.e.
\begin{equation}
  \vc_{t_{k + 1}} = \vc_{t_{k}} + \int_{t_{k}}^{t_{k + 1}} -\frac{1}{t} \mAhippo \vc_{t} + \frac{1}{t} \mBhippo f(t)\,\diff{t}.
\end{equation}
For discrete observations $f(t_{1}), f(t_{2}), \ldots$, we need to approximate the integral on the right-hand side in a way that only requires $f(t_{k})$ and $f(t_{k + 1})$.

\citep{gu2020hippo} describe three approximations for the integral, \emph{forward Euler}, \emph{backward Euler} and the \emph{trapezoidal rule} of which they use the latter for their experiments.
Forward Euler approximates the value of the integrand with its value at the lower integration bound, i.e.\ at $t_{k}$, and gives
\begin{equation}
  \vc_{t_{k + 1}} \approx \vc_{t_{k}} + \underbrace{(t_{k + 1} - t_{k})}_{\eqcolon \Delta t}\,\bigg( -\frac{1}{t_{k}} \mAhippo \vc_{t_{k}} + \frac{1}{t_{k}} \mBhippo f(t_{k}) \bigg) = \bigg( \mI - \frac{\Delta t}{t_{k}}\mAhippo \bigg)\,\vc_{t_{k}} + \frac{\Delta t}{t_{k}} \mBhippo f(t_{k}). \label{eq:forward-euler}
\end{equation}
Backward Euler is the other extreme that approximates the integrand with its value at $t_{k + 1}$, i.e.
\begin{equation}
  \vc_{t_{k + 1}} \approx \vc_{t_{k}} + \Delta t\,\bigg( -\frac{1}{t_{k + 1}} \mAhippo \vc_{t_{k + 1}} + \frac{1}{t_{k + 1}} \mBhippo f(t_{k + 1}) \bigg).
\end{equation}
Solving for $\vc_{t_{k + 1}}$ gives us
\begin{equation}
  \vc_{t_{k + 1}} \approx {\bigg( \mI + \frac{\Delta t}{t_{k + 1}}\mAhippo \bigg)}^{\!\shortminus{}1} \bigg[ \vc_{t_{k}} + \frac{\Delta t}{t_{k + 1}} \mBhippo f(t_{k + 1}) \bigg]. \label{eq:backward-euler}
\end{equation}
For the trapezoidal rule, we approximate the integrand as the average of its values at $t_{k}$ and $t_{k + 1}$, i.e.
\begin{equation}
  \vc_{t_{k + 1}} \approx \vc_{t_{k}} + \Delta t\,\frac{1}{2}\,\bigg[ \bigg( -\frac{1}{t_{k}} \mAhippo \vc_{t_{k}} + \frac{1}{t_{k}} \mBhippo f(t_{k}) \bigg) + \bigg( -\frac{1}{t_{k + 1}} \mAhippo \vc_{t_{k + 1}} + \frac{1}{t_{k + 1}} \mBhippo f(t_{k + 1}) \bigg) \bigg].
\end{equation}
After simplifying and solving for $\vc_{t_{k + 1}}$, this becomes
\begin{equation}
  \vc_{t_{k + 1}} \approx {\bigg( \mI + \frac{\Delta t}{2t_{k + 1}}\mAhippo \bigg)}^{\!\shortminus{}1} \bigg[ \bigg( \mI - \frac{\Delta t}{2t_{k}}\mAhippo \bigg)\vc_{t_{k}} + \frac{\Delta t}{2}\mBhippo \bigg( \frac{1}{t_{k}}f(t_k) + \frac{1}{t_{k + 1}}f(t_{k + 1}) \bigg) \bigg]. \label{eq:trapezoidal}
\end{equation}

Note that \citep{gu2020hippo} approximate \cref{eq:trapezoidal} as
\begin{equation}
  \vc_{t_{k + 1}} \approx {\bigg( \mI + \frac{\Delta t}{2t_{k+1}}\mAhippo \bigg)}^{\!\shortminus{}1} \bigg[ \bigg( \mI - \frac{\Delta t}{2t_{k+1}}\mAhippo \bigg)\vc_{t_{k}} + \frac{\Delta t}{t_{k+1}}\mBhippo f(t_{k + 1}) \bigg], \label{eq:trapezoidal-lssl}
\end{equation}
i.e.\ they approximate $\nicefrac{1}{t_{k}}f(t_k) + \nicefrac{1}{t_{k + 1}}f(t_{k + 1})$ as $\nicefrac{2}{t_{k + 1}}f(t_{k + 1})$.
This, in effect, pretends that the dynamics in \cref{eq:hippo-dynamics} would be constant in time and is required by \lssl{} for the numerical stability of the accelerated evaluation via convolution with a Krylov kernel.
In this work, we discretize the \lssl{} baseline with \cref{eq:trapezoidal-lssl}, too.
Note that for our model a closed-form solution in terms of the matrix exponential exists, which we use instead of a numerical approximation.

Each of \cref{eq:forward-euler,eq:backward-euler,eq:trapezoidal,eq:trapezoidal-lssl} can be put into the form
\begin{equation}
  \vc_{t_{k + 1}} = \bar{\mA}_{\mathrm{H},t_{k}}\vc_{t_{k}} + \bar{\mB}_{\mathrm{H},t_{k}}f(t_{k}) + \bar{\mB}_{\mathrm{H},t_{k + 1}}f(t_{k + 1})
\end{equation}
by collecting terms appropriately.

\section{Discretizations Visualized}\label{sec:discretization-visual}

\begin{figure}[H]
  \centering
  \includegraphics{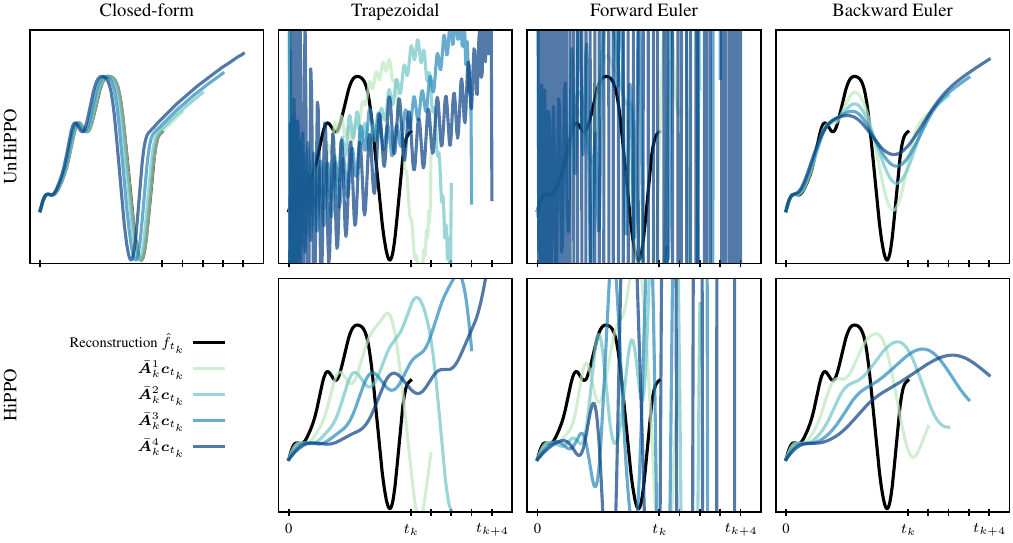}
  \caption{Visualization of repeated application of \hippo{} and \unhippo{} matrices with various discretization methods through the polynomial representation $\fhat$ of $\vc$. Magnitude of the plotted curves correlates with $\| \bar{\mA}_{k}^{i}\vc_{t_{k}} \|$.}\label{fig:discretizations}
\end{figure}

Recurrent roll-out of an \ssm{} or the construction of the Krylov kernel equate to repeated matrix multiplication with the \hippo{} or \unhippo{} matrix or taking powers thereof.
Therefore, successful training and evaluation on long sequences requires the discretization to be chosen so that matrix powers do not diverge.
\cref{fig:discretizations} shows the effect of repeated applications of \hippo{} and \unhippo{} matrices discretized with the closed-form solution and the approximations considered by \citep{gu2021combining}.
The \hippo{} matrix is stable under the backward discretizations and has a negligible degree of divergence with the trapezoidal rule.
The \unhippo{} matrix cannot be stably discretized with the trapezoidal rule, but its closed-form discretization is stable and retains the information encoded in the shape of the reconstruction $\fhat$ almost perfectly.

\section{Experiment Details}\label{sec:experiment-details}

We use the parameters listed in \cref{tab:lssl-parameters} for all models.
For the range of $t_{k}$ in the initialization of \lssl{}, we set $t_{\mathrm{min}} = 10$ and $t_{\mathrm{max}} = 1000$ to cover a range of time scales.

\begin{table}[H]
  \centering
  \caption{Hyperparameters of the LSSL architecture used for the SC10 experiments.}\label{tab:lssl-parameters}
  \begin{tabular}{cc}
    \toprule
    Parameter             & Value     \\
    \midrule
    Layers                & 4         \\
    \ncoefs{}             & 128       \\
    Linear Embedding Size & 128       \\
    Latent Channels       & 4         \\
    Dropout               & 0.1       \\
    UnHiPPO $\sigma^{2}$  & $10^{10}$ \\
    Training Steps        & 100000    \\
    Batch Size            & 16        \\
    \bottomrule
  \end{tabular}
\end{table}

\end{document}